\documentclass[letterpaper]{article} 
\usepackage{aaai2027}
\usepackage[hyphens]{url}  
\usepackage{graphicx} 
\usepackage{natbib}  
\usepackage{caption} 

\usepackage{xcolor}
\usepackage{algorithm}
\usepackage{algorithmic}
\usepackage{amsfonts}
\usepackage{amsmath}
\usepackage{amsthm}

\usepackage{newfloat}
\usepackage{listings}

\usepackage{xcolor}

\DeclareCaptionStyle{ruled}{labelfont=normalfont,labelsep=colon,strut=off} 
\floatstyle{ruled}
\newfloat{listing}{tb}{lst}{}
\floatname{listing}{Listing}

\usepackage{booktabs}
\usepackage{algorithm}
\usepackage{algorithmic}
\usepackage{booktabs}
\usepackage{amsmath}
\usepackage{amssymb}
\usepackage{multirow}
\title{Gradient-Guided Decoupled Adaptation for Geospatial Vision-Language Models}
\author{
    Dongdong Wang\textsuperscript{\rm 1}\corresponding,
    Deepak Balakrishnan\textsuperscript{\rm 1},
    Ravi Srinivasan\textsuperscript{\rm 1},
    Shenhao Wang\textsuperscript{\rm 1}\corresponding
}

\affiliations{
    \textsuperscript{\rm 1}University of Florida\\
    Gainesville, FL, USA\\
    \{dongdongwang, deepakbd, sravi, shenhaowang\}@ufl.edu
}

\makeatletter
\renewcommand{\copyright@on}{}
\makeatother

\begin{document}

\maketitle

\begin{abstract}

Existing geospatial vision-language models (Geo-VLMs) typically optimize diverse geospatial tasks through a unified multi-task adaptation paradigm without explicitly accounting for the heterogeneous optimization characteristics. Our empirical observations reveal heterogeneous gradient characteristics across tasks, including vision-language differences, intra-branch gradient relationships, and task interference, which hinder effective multi-task optimization. Motivated by these observations, we propose Gradient-Guided Decoupled Adaptation (G$^2$DA), a gradient-aware optimization framework for multi-task Geo-VLM learning. G$^2$DA first partitions tasks into vision- and language-centric groups through gradient-guided cross-modal decoupling. It then constructs modality-specific curricula based on task gradient similarity and employs bidirectional rehearsal to mitigate the recency effects introduced by sequential optimization. We evaluate G$^2$DA on three Geo-VLM benchmarks using six InternVL3 and Qwen3.5-VL variants, along with GeoChat and GeoLLaVA. Across all 24 benchmark--model combinations, G$^2$DA consistently outperforms representative baselines, improving over the strongest competitor by 3.08, 4.30, and 2.81 percentage points on UrBench-MCQ, XLRS-Bench-Lite, and VRS-Bench-VQA, respectively. These results demonstrate the effectiveness of gradient-guided task organization for Geo-VLM adaptation.

\end{abstract}

\section{Introduction}
Recent vision-language models (VLMs) have achieved strong performance across a wide range of multimodal reasoning tasks. Building on these advances, geospatial vision-language models (Geo-VLMs) have emerged as a powerful framework for geospatial intelligence by jointly modeling multi-view geospatial imagery and natural language \cite{mai2024opportunities}. To support diverse reasoning capabilities within a single model, including visual perception, object grounding, spatial reasoning, retrieval, and quantitative analysis, existing approaches predominantly adopt unified multi-task learning, where tasks with substantially different objectives are jointly optimized using shared model parameters \cite{kuckreja2024geochat, feng2025urbanllava, wang2026geollava}. 

While unified multi-task learning facilitates knowledge sharing across tasks, it jointly optimizes heterogeneous tasks using shared model parameters, implicitly assuming compatible optimization objectives. However, geospatial tasks exhibit substantially different optimization characteristics due to their varying reliance on visual perception and language reasoning \cite{zhou2025urbench}. Consequently, they generate heterogeneous optimization signals, and jointly optimizing them within a shared parameter space often introduces conflicting gradients that hinder effective multi-task adaptation \cite{jeong2025selective}.

Existing multi-task optimization methods mitigate gradient conflicts through projection~\cite{yu2020gradient}, reweighting~\cite{chen2018gradnorm}, or balancing~\cite{liu2023famo}, yet still optimize all tasks within a unified parameter space. This assumption is particularly challenging for Geo-VLMs, where tasks exhibit heterogeneous optimization demands across vision and language encoders due to their varying reliance on visual perception and language reasoning. Consequently, optimization strategies designed for a unified parameter space may fail to fully exploit these cross-modal characteristics. Our empirical gradient analysis reveals substantial gradient conflicts across the task set, distinct optimization preferences between the vision and language components, and persistent intra-encoder interference within each component.


Motivated by these observations, we propose G$^2$DA, a gradient-guided optimization framework for Geo-VLM adaptation. Unlike prior methods that optimize all tasks within a unified parameter space, G$^2$DA organizes multi-task optimization according to heterogeneous cross-modal gradient characteristics through gradient-guided decoupling, modality-specific curriculum learning, and bidirectional rehearsal. Our contributions are summarized as follows:

\begin{itemize}
    \item We present a gradient-centric analysis of multi-task optimization in Geo-VLMs. Our empirical study reveals hierarchical optimization heterogeneity, including cross-task gradient conflicts, distinct optimization preferences between vision and language components, and persistent intra-component task interference.

    \item Given the empirical analyses, we design G$^2$DA, a unified gradient-guided optimization framework that integrates cross-modal optimization decoupling, modality-specific curriculum learning, and cross-task rehearsal to address optimization heterogeneity in Geo-VLM learning.

    \item Extensive experiments on multiple Geo-VLM benchmarks and diverse model families demonstrate that G$^2$DA consistently outperforms representative baselines, establishing gradient-guided optimization as an effective paradigm for adapting Geo-VLMs.
\end{itemize}
\section{Related Work}

\subsection{Geospatial Vision-Language Models}

Recent advances in VLMs~\cite{bai2023qwen, chen2024intern} have enabled the adaptation of general-purpose VLMs to geospatial domains through instruction tuning~\cite{liu2023visual}, leading to models such as GeoChat~\cite{kuckreja2024geochat} and GeoLLaVA~\cite{wang2026geollava}. At the same time, geospatial benchmarks including UrBench~\cite{zhou2025urbench}, XLRS-Bench~\cite{wang2025XLRS}, and VRS-Bench~\cite{li2024vrsbench} have accelerated Geo-VLM development by supporting unified training and evaluation across diverse reasoning tasks. Despite their architectural differences, existing Geo-VLMs are typically adapted through unified multi-task learning, jointly optimizing heterogeneous tasks under a shared objective without considering their optimization compatibility~\cite{kuckreja2024geochat,feng2025urbanllava,wang2026geollava}. As task diversity continues to increase, this unified optimization strategy can introduce gradient interference, limiting the efficiency and effectiveness of multi-task adaptation.

\subsection{Gradient Conflict in Multi-task Learning}

Multi-task learning improves generalization by jointly optimizing multiple tasks within a shared model. Existing optimization methods primarily mitigate task interference caused by conflicting gradients. PCGrad~\cite{yu2020gradient} alleviates conflicts through gradient projection, while CAGrad~\cite{liu2021conflict} computes conflict-aware gradient updates via adaptive gradient combination. GradNorm~\cite{chen2018gradnorm} and FAMO~\cite{liu2023famo} balance task optimization through adaptive gradient weighting. More recently, Selective Group Update (STGU)~\cite{jeong2025selective} shifts the focus from gradient modification to task scheduling, demonstrating improved optimization for heterogeneous multi-task learning. However, these methods, including STGU, still optimize heterogeneous tasks within a unified parameter space, overlooking the distinct optimization characteristics of the vision and language components in Geo-VLMs. Consequently, they are not designed to exploit task-specific cross-modal optimization preferences that naturally arise in Geo-VLM adaptation.

\subsection{Curriculum and Rehearsal Learning}

Curriculum learning improves optimization by organizing training samples or tasks into a structured learning sequence rather than random presentation~\cite{bengio2009curriculum}, and has been extended to multi-task learning through task-level scheduling strategies that improve optimization efficiency~\cite{wang2021survey}. However, sequential curricula may introduce optimization imbalance across tasks as training progresses and earlier tasks are no longer revisited. Indeed, one-pass curriculum scheduling has been linked to continual learning, where stage-wise optimization may lead to catastrophic forgetting~\cite{wang2021survey}. Rehearsal-based learning mitigates this issue by revisiting previously learned samples or tasks during optimization, and has proven particularly effective for heterogeneous tasks with low similarity~\cite{deng2025unlocking}. Nevertheless, existing rehearsal methods are primarily designed for continual learning, while their application to mitigating optimization imbalance in sequential curriculum learning for multi-task Geo-VLM adaptation remains unexplored.
\section{Method}
\label{sec:method}
\subsection{Problem Formulation}

Let $\mathcal{T}=\{T_1,T_2,\ldots,T_N\}$ denote a set of Geo-VLM tasks. Each task $T_i$ is associated with a dataset $\mathcal{D}_i=\{(\mathbf{x},\mathbf{q},\mathbf{y})\}$, where $\mathbf{x}$ is an image, $\mathbf{q}$ is a language query, and $\mathbf{y}$ is the supervision. A unified Geo-VLM $f(\cdot;\Theta)$ is optimized over all tasks with the objective
\begin{equation}
\mathcal{L}(\Theta)=
\sum_{i=1}^{N}\lambda_i\mathcal{L}_i(\Theta),
\end{equation}
where $\mathcal{L}_i$ is the task-specific loss and $\lambda_i$ balances its contribution to the overall objective.

Unlike conventional multi-task learning, which primarily aims to improve the converged solution, we study a fixed-budget optimization setting. Let $B$ denote the total number of parameter updates available during training. Under the same initialization $\Theta_0$ and optimization budget $B$, the final optimization outcome depends on how heterogeneous task data are presented throughout training. We represent the entire training schedule as

\begin{equation}
\Pi=\left\{D^{(1)},D^{(2)},\ldots,D^{(B)}\right\},
\end{equation}

where each mini-batch
$D^{(t)}\in\bigcup_{i=1}^{N}\mathcal D_i$
is sampled from one of the task datasets. Different schedules $\Pi$ induce different optimization trajectories and therefore different model parameters after exactly $B$ updates, denoted by $\Theta_B^{\Pi}$.

Our objective is to design an effective training schedule that improves the joint optimization objective under the same optimization budget:

\begin{equation}
\mathcal L(\Theta_B^{\Pi})
=
\sum_{i=1}^{N}
\lambda_i
\mathcal L_i(\Theta_B^{\Pi}),
\qquad
\text{s.t.}\quad
|\Pi|=B.
\end{equation}

This formulation raises a fundamental question: \emph{How should heterogeneous Geo-VLM tasks be organized under a fixed optimization budget?} To answer this question, we analyze task gradients to characterize optimization heterogeneity and use these observations to guide the design of G$^2$DA.


\subsection{Gradient Characterization}

\begin{figure}[h]
    \centering
    \includegraphics[width=0.99\linewidth]{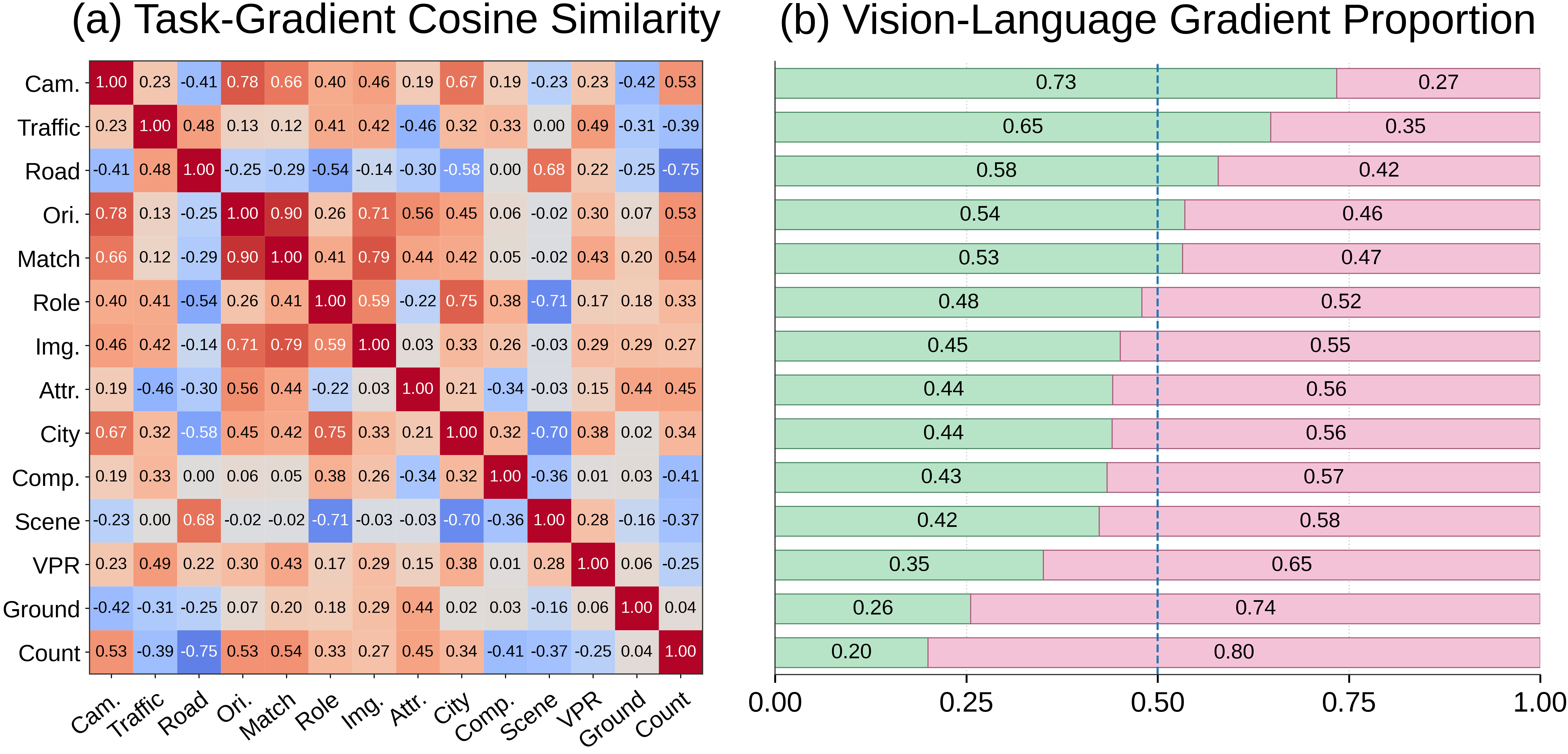}
    \caption{Task-gradient analysis on UrBench-MCQ. (a) Pairwise task-gradient cosine similarity matrix. Positive (red) and negative (blue) values indicate aligned and conflicting optimization directions. (b) Normalized proportions of the total gradient norm contributed by the vision (green) and language (pink) encoders for each task.}
    \label{fig:gradient_similarity}
\end{figure}

To better understand the optimization challenges under a fixed update budget, we characterize heterogeneous geospatial vision-language tasks through gradient analysis. Specifically, for each task, we compute the gradient with respect to the trainable parameters and use the gradient cosine similarity as a proxy to quantify pairwise optimization alignment.
\begin{equation}
S_{ij}=
\frac{\mathbf g_i^\top \mathbf g_j}
{\|\mathbf g_i\|_2\|\mathbf g_j\|_2},
\label{eq:similarity}
\end{equation}
where $\mathbf g_i$ and $\mathbf g_j$ denote the task gradients of tasks $i$ and $j$, respectively. 

To characterize each task's cross-modal optimization preference, we further compute its vision gradient proportion as another proxy,
\begin{equation}
r_i=
\frac{\|\mathbf g_i^{v}\|_2}
{\|\mathbf g_i^{v}\|_2+\|\mathbf g_i^{l}\|_2},
\label{eq:proportion}
\end{equation}
where $\mathbf g_i^{v}$ and $\mathbf g_i^{l}$ represent the gradients on the vision and language encoders, respectively. With this proxy, a larger $r_i$ indicates stronger optimization demand on the vision encoder, while a smaller value indicates greater reliance on the language encoder. Together, the gradient cosine similarity and gradient proportion characterize inter-task optimization compatibility and cross-modal optimization preference, respectively. 

We first use UrBench~\cite{zhou2025urbench}, where all questions are reformulated into multiple-choice format (UrBench-MCQ), as a pilot benchmark to empirically analyze the gradient behavior of heterogeneous Geo-VLM tasks using Qwen3.5-VL-2B. The resulting analyses reveal three observations that motivate the design of G$^2$DA. Corresponding results on additional benchmarks are included in the supplementary material.

\textbf{Observation 1: Inter-task gradient conflict.} Different Geo-VLM tasks often induce substantially
different optimization directions. As shown by the pairwise gradient cosine
similarity matrices, some task pairs exhibit highly aligned gradients,
whereas others are nearly orthogonal or negatively correlated. A negative
cosine similarity indicates that decreasing the loss of one task may increase
the loss of another under the same parameter update. Therefore, uniformly
mixing heterogeneous tasks in conventional multi-task training introduces
optimization interference and reduces optimization efficiency.

\textbf{Observation 2: Cross-modal optimization imbalance.} Different tasks exhibit substantially different vision–language gradient proportions: some are predominantly optimized through visual representations, whereas others rely primarily on language reasoning, with many requiring balanced optimization across both modalities. This heterogeneous
optimization demand suggests that applying identical parameter updates across
both encoders is inefficient, motivating modality-aware optimization.

\textbf{Observation 3: Intra-encoder gradient heterogeneity.} Even after separating tasks according to their cross-modal optimization
characteristics, substantial gradient heterogeneity remains within each
encoder. Pairwise gradient cosine similarities reveal that tasks optimized on
the same encoder still exhibit diverse optimization directions, indicating
that encoder decoupling alone cannot fully eliminate optimization
interference. This observation motivates gradient-aware curriculum scheduling
within each encoder.

\begin{figure}[h]
    \centering
    \includegraphics[width=1\linewidth]{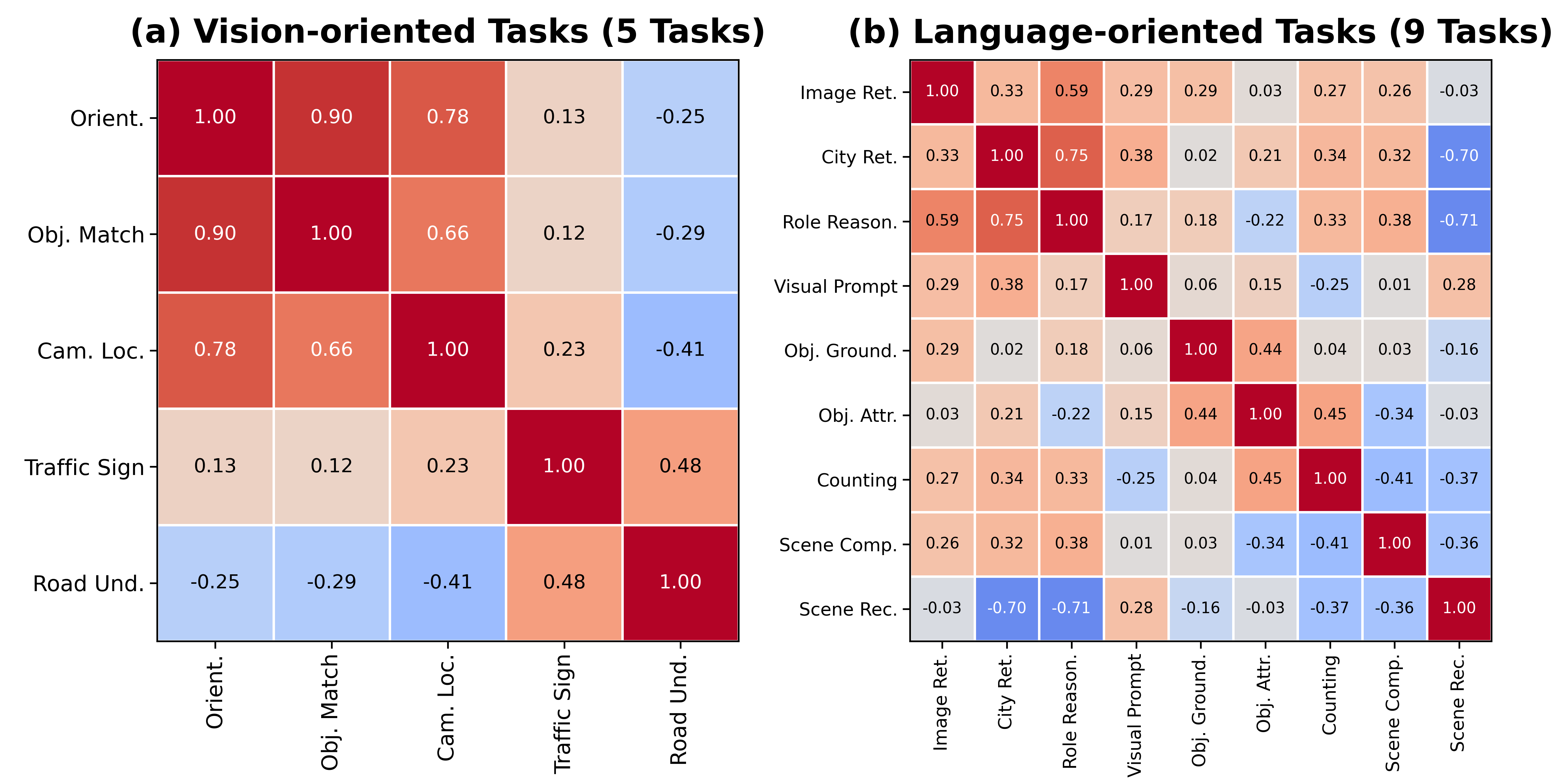}
    \caption{Task-level gradient cosine similarity after cross-modal decoupling and curriculum ordering on UrBench-MCQ.}
    \label{fig:local_matrices}
\end{figure}

\subsection{Optimization}

\begin{figure*}
    \centering
    \includegraphics[width=0.9\linewidth]{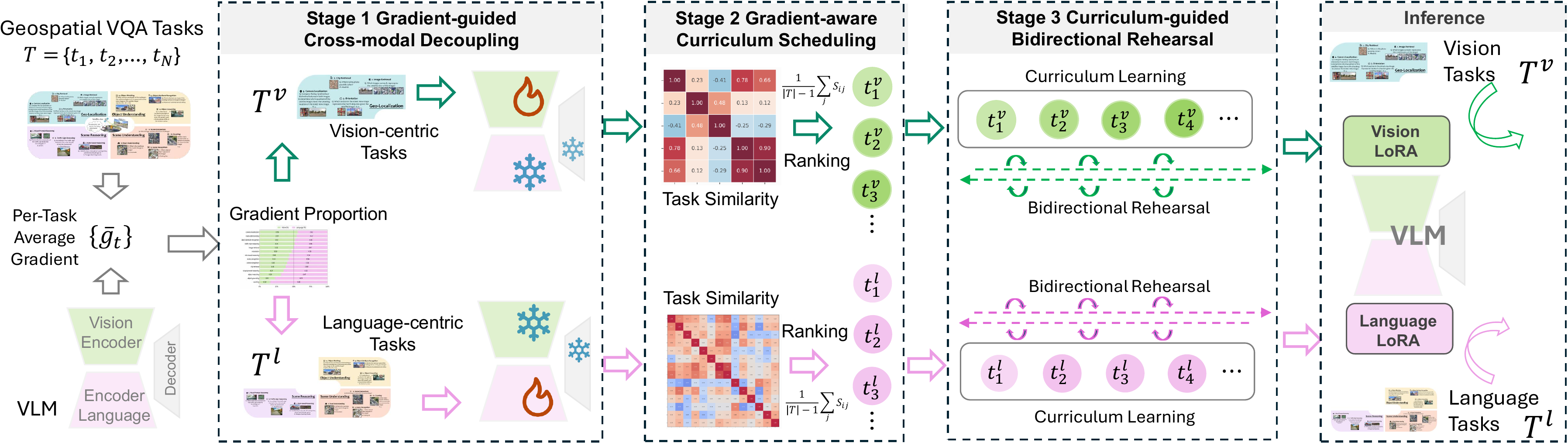}
    \caption{Overview of the proposed G$^2$DA framework, consisting of gradient-guided cross-modal decoupling, gradient-aware curriculum scheduling, and bidirectional rehearsal.}
    \label{fig:g2da}
\end{figure*}


Motivated by the observations above, we propose G$^2$DA, a unified optimization framework for fixed-budget Geo-VLM adaptation. Unlike conventional multi-task learning, Geo-VLM tasks exhibit heterogeneous optimization preferences across vision and language encoders due to their varying reliance on visual perception and language reasoning. Therefore, rather than organizing tasks solely according to pairwise gradient similarity, G$^2$DA first decouples tasks based on their cross-modal optimization preferences and then models task compatibility within each modality through gradient-aware curriculum scheduling, followed by bidirectional rehearsal for knowledge preservation. The overall framework is illustrated in Figure~\ref{fig:g2da}.

\textbf{Gradient-guided cross-modal decoupling.}
Motivated by Observation~2, we use the vision gradient proportion $r_i$ in Eq.~\ref{eq:proportion} as a proxy to characterize each task's relative preference for the vision and language encoders. Based on the estimated gradient proportion $r_i$, tasks are divided into two groups: \emph{vision-centric} and \emph{language-centric} tasks. Grouping tasks with similar cross-modal preferences reduces interference caused by heterogeneous encoder demands and allows the optimization budget to be allocated more effectively. As shown in Figure~\ref{fig:local_matrices}, cross-modal decoupling increases local gradient cosine similarity, indicating improved optimization compatibility within each branch. However, considerable intra-branch gradient conflicts remain, motivating the branch-specific curriculum learning introduced in the next stage.

\textbf{Gradient-aware curriculum scheduling.}
Motivated by Observation~3, we compute gradient cosine similarities within each modality-centric optimization group to characterize task compatibility. Specifically, for task $t_i$, the curriculum score is defined as
\begin{equation}
\phi_i^e =
\frac{1}{|\mathcal T^e|-1}
\sum_{\substack{j\in\mathcal T^e\\j\neq i}}
\frac{
(\bar{\mathbf g}_i^{\,e})^\top\bar{\mathbf g}_j^{\,e}
}{
\|\bar{\mathbf g}_i^{\,e}\|_2
\|\bar{\mathbf g}_j^{\,e}\|_2
},
\end{equation}
where $\mathcal{T}^e$ denotes one of the vision- or language-centric task groups. Tasks with higher average compatibility are optimized earlier, yielding a curriculum that progresses from broadly aligned to relatively conflicting tasks. By constructing the curriculum within each modality-centric group, the proposed strategy replaces random shuffling with gradient-aware sequential training, improving optimization compatibility between consecutive tasks. However, this sequential optimization also introduces recency effects and catastrophic forgetting \cite{kemker2018measuring}, motivating the bidirectional rehearsal strategy described next.

\textbf{Bidirectional rehearsal for knowledge preservation.}
Once the curriculum is established, tasks are optimized sequentially according to the curriculum order. Such sequential optimization inevitably introduces a recency effect, where knowledge acquired in earlier stages is gradually overwritten by later stages. To alleviate this issue, we optimize each task group in both the ascending curriculum order and its reverse, allowing every curriculum stage to be optimized once in descending compatibility order and once in the reverse order. Meanwhile, for each curriculum direction, the effective training set of stage $k$ is defined as
\begin{equation}
\tilde{\mathcal D}_k
=
\mathcal D_k
\cup
\mathcal M_k,
\end{equation}
where $\mathcal D_k$ denotes the current curriculum stage and
\begin{equation}
\mathcal M_k
\sim
\mathrm{Uniform}(\mathcal D_k^{c}),
\qquad
|\mathcal M_k|
=
|\mathcal D_k|,
\end{equation}
is a rehearsal set uniformly sampled from the complementary curriculum stages $\mathcal D_k^{c}$ such that the total number of rehearsal samples equals the size of the current stage. During optimization, mini-batches are uniformly sampled from the effective training set $\tilde{\mathcal D}_k$. By traversing the curriculum in both ascending and descending similarity orders while jointly optimizing the current-stage and rehearsal samples, the proposed bidirectional rehearsal preserves knowledge across the entire curriculum, mitigates optimization drift, and improves balanced performance across heterogeneous tasks.

\subsection{G$^2$DA Framework}

Algorithm~\ref{alg:optimization} summarizes the optimization procedure of G$^2$DA. Given the task-level average gradients $\{\bar{\mathbf g}_t^v,\bar{\mathbf g}_t^l\}_{t=1}^{T}$, G$^2$DA first computes the gradient proportion $r_t$ for each task and partitions the task set into $\mathcal T^v$ and $\mathcal T^l$. For each modality $e\in\{v,l\}$, pairwise gradient cosine similarities $S_{ij}^e$ are computed to obtain the curriculum scores $\phi_i^e$, from which the curriculum order $\Pi^e$ is constructed. Finally, for each curriculum order and its reverse, rehearsal samples $\mathcal M_k$ are uniformly drawn from the remaining datasets to form the augmented training set $\tilde{\mathcal D}_k$, which is used to optimize the corresponding parameters $\Theta^e$.
\begin{algorithm}[h]
\caption{G$^2$DA Optimization}
\label{alg:optimization}
\begin{algorithmic}[1]
\REQUIRE Task datasets $\{\mathcal D_t\}_{t=1}^{T}$, parameters
$\Theta=\{\Theta^v,\Theta^l\}$
\ENSURE Optimized parameters $\Theta$

\STATE Compute task-level average gradients
$\{\bar{\mathbf g}_t^v,\bar{\mathbf g}_t^l\}_{t=1}^{T}$
\vspace{0.6em}
\STATE \textit{Stage 1: Cross-modal decoupling}
\FOR{$t=1,\ldots,T$}
    \STATE $r_t \leftarrow
    \frac{\|\bar{\mathbf g}_t^v\|_2}
    {\|\bar{\mathbf g}_t^v\|_2+\|\bar{\mathbf g}_t^l\|_2}$
\ENDFOR
\STATE $\mathcal T^v\leftarrow\{t:r_t>0.5\}$,\quad
$\mathcal T^l\leftarrow\{t:r_t\leq0.5\}$
\vspace{0.6em}
\STATE \textit{Stage 2: Gradient-aware curriculum}
\FOR{$e\in\{v,l\}$}
    \STATE $S_{ij}^e\leftarrow
    \cos(\bar{\mathbf g}_i^e,\bar{\mathbf g}_j^e)$,
    \quad $i,j\in\mathcal T^e$
    \STATE $\phi_i^e\leftarrow
    \frac{1}{|\mathcal T^e|-1}\sum_{j\neq i}S_{ij}^e$
    \STATE $\Pi^e\leftarrow
    \mathrm{Sort}_{\downarrow}
    (\{\phi_i^e\}_{i\in\mathcal T^e})$
\ENDFOR
\vspace{0.6em}
\STATE \textit{Stage 3: Bidirectional rehearsal}
\FOR{$e\in\{v,l\}$}
    \FOR{$\pi\in\{\Pi^e,\mathrm{Reverse}(\Pi^e)\}$}
        \FOR{$\mathcal D_k\in\pi$}
            \STATE $\mathcal D_k^c\leftarrow
            \bigcup_{\mathcal D_i\in\pi,\,i\neq k}\mathcal D_i$
            \STATE $\mathcal M_k\sim
            \mathrm{Uniform}(\mathcal D_k^c)$,
            \quad $|\mathcal M_k|=|\mathcal D_k|$
            \STATE $\tilde{\mathcal D}_k\leftarrow
            \mathcal D_k\cup\mathcal M_k$
            \STATE Optimize $\Theta^e$ on $\tilde{\mathcal D}_k$
        \ENDFOR
    \ENDFOR
\ENDFOR

\RETURN $\Theta$
\end{algorithmic}
\end{algorithm}

\section{Experiment}

\subsection{Experimental Setup}

\paragraph{Benchmarks and evaluation.}
We evaluate G$^2$DA on three Geo-VLM benchmarks: UrBench-MCQ, XLRS-Bench-Lite, and the VQA subset of VRS-Bench (VRS-Bench-VQA). All VQA tasks are reformulated as multiple-choice questions with images resized to $1024\times1024$. Performance is measured by accuracy (\%) averaged across all tasks. All experiments are repeated with three random seeds, and the mean and standard deviation are reported. Additional benchmark details are provided in the supplementary material.

\paragraph{Models and baselines.}
We evaluate G$^2$DA on two general-purpose VLM families, InternVL3 and Qwen3.5-VL, across multiple model scales, together with two representative Geo-VLMs, GeoChat and GeoLLaVA. We compare against representative multi-task optimization methods, including Joint training, GradNorm~\cite{chen2018gradnorm}, PCGrad~\cite{yu2020gradient}, CAGrad~\cite{liu2021conflict}, FAMO~\cite{liu2023famo}, and STGU~\cite{jeong2025selective}, covering gradient balancing, conflict mitigation, adaptive multi-objective optimization, and task scheduling. All methods are evaluated under identical optimization budgets.

\paragraph{Implementation details.}
Experiments are conducted on four NVIDIA B200 GPUs (180\,GB each) using \texttt{bfloat16} precision and the AdamW optimizer. The batch size is 1 with gradient accumulation over 8 steps, and the generation temperature is fixed at 1. Baselines are trained for 8 epochs, while G$^2$DA is trained for 2 epochs. Since each epoch of G$^2$DA performs both forward and reverse curricula with bidirectional rehearsal, the total number of optimization steps is comparable across methods. The learning rate and weight decay are $5\times10^{-6}$ and $1\times10^{-6}$, respectively. LoRA adapters use rank $r=8$, scaling factor $\alpha=16$, and dropout $0.1$.

\subsection{Results}

Table~\ref{tab:benchmark} summarizes the overall performance. G$^2$DA achieves the best results in all 24 benchmark-model combinations, indicating consistent performance across diverse architectures and model scales. Compared with the strongest competing baseline in each setting, G$^2$DA still improves performance. On UrBench-MCQ, G$^2$DA improves performance by 3.08 pp on average, with gains ranging from 1.0 to 7.0 pp. Existing gradient manipulation methods, such as PCGrad and CAGrad, often perform comparably to or even worse than joint optimization, while STGU is generally the strongest baseline. On the more challenging XLRS-Bench-Lite, the advantage of G$^2$DA becomes more pronounced. G$^2$DA improves upon the strongest baseline by an average of 4.30 pp, with improvements reaching 8.0 pp on Qwen3.5-VL-27B and 7.0 pp on InternVL3-14B. On VRS-Bench-VQA, G$^2$DA achieves an average improvement of 2.81 pp over the strongest baseline, with gains of up to 5.0 pp. Table~\ref{tab:benchmark} also evaluates G$^2$DA on two representative Geo-VLMs, GeoChat and GeoLLaVA. Compared with the strongest competing baseline, G$^2$DA improves performance by 2-5 pp, demonstrating that the proposed optimization framework is not limited to general-purpose VLMs but also transfers effectively to domain-specific Geo-VLMs.

\begin{table*}[t]
\centering
\caption{Performance (\%) over three independent runs. Best results are shown in \textbf{bold}, and second-best are \underline{underlined}.}
\label{tab:benchmark}
\footnotesize
\setlength{\tabcolsep}{4.2pt}
\renewcommand{\arraystretch}{1.05}
\begin{tabular}{llccccccc}
\toprule
\multirow{2}{*}{Benchmark} &
\multirow{2}{*}{Model} &
\multicolumn{7}{c}{Optimization Method} \\
\cmidrule(lr){3-9}
&
&
Joint &
PCGrad &
GradNorm &
STGU &
CAGrad &
FAMO &
\textbf{G$^2$DA} \\
\midrule

\multirow{8}{*}{UrBench-MCQ}

& GeoChat-7B
& $51.1_{\pm0.2}$ & $48.5_{\pm0.2}$ & $49.9_{\pm0.2}$ & \underline{$51.2_{\pm0.3}$} & $48.3_{\pm0.2}$ & $50.2_{\pm0.3}$ & $\mathbf{54.1_{\pm0.2}}$ \\

& GeoLLaVA-8B
& \underline{$51.7_{\pm0.2}$} & $48.5_{\pm0.2}$ & $50.4_{\pm0.3}$ & $50.3_{\pm0.2}$ & $48.6_{\pm0.2}$ & $50.3_{\pm0.2}$ & $\mathbf{54.8_{\pm0.2}}$ \\
\cmidrule(lr){2-9}

& InternVL3-2B
& $44.2_{\pm0.2}$ & $39.2_{\pm0.1}$ & $43.4_{\pm0.3}$ & \underline{$45.8_{\pm0.2}$} & $39.9_{\pm0.2}$ & $43.2_{\pm0.3}$ & $\mathbf{48.8_{\pm0.3}}$ \\

& InternVL3-8B
& $51.9_{\pm0.1}$ & $48.1_{\pm0.4}$ & $51.9_{\pm0.4}$ & \underline{$54.9_{\pm0.3}$} & $48.1_{\pm0.2}$ & $51.8_{\pm0.2}$ & $\mathbf{57.8_{\pm0.1}}$ \\

& InternVL3-14B
& $61.1_{\pm0.2}$ & $53.8_{\pm0.2}$ & $61.1_{\pm0.2}$ & \underline{$63.8_{\pm0.2}$} & $54.1_{\pm0.1}$ & $61.4_{\pm0.2}$ & $\mathbf{66.8_{\pm0.2}}$ \\

& Qwen3.5-VL-2B
& $44.8_{\pm0.1}$ & $40.9_{\pm0.3}$ & $45.0_{\pm0.3}$ & \underline{$46.1_{\pm0.2}$} & $41.1_{\pm0.3}$ & \underline{$46.1_{\pm0.2}$} & $\mathbf{49.0_{\pm0.3}}$ \\

& Qwen3.5-VL-8B
& $55.0_{\pm0.3}$ & $52.0_{\pm0.3}$ & \underline{$56.0_{\pm0.3}$} & \underline{$56.0_{\pm0.4}$} & $52.0_{\pm0.3}$ & $55.0_{\pm0.1}$ & $\mathbf{59.0_{\pm0.3}}$ \\

& Qwen3.5-VL-27B
& $64.8_{\pm0.3}$ & $64.1_{\pm0.3}$ & $60.2_{\pm0.3}$ & \underline{$65.1_{\pm0.3}$} & $60.9_{\pm0.3}$ & $64.1_{\pm0.2}$ & $\mathbf{68.7_{\pm0.2}}$ \\

\midrule

\multirow{8}{*}{XLRS-Bench-Lite}

& GeoChat-7B
& $38.4_{\pm0.2}$ & $35.3_{\pm0.2}$ & $37.4_{\pm0.2}$ & \underline{$39.0_{\pm0.2}$} & $35.3_{\pm0.2}$ & $38.5_{\pm0.2}$ & $\mathbf{41.0_{\pm0.2}}$ \\

& GeoLLaVA-8B
& \underline{$43.8_{\pm0.2}$} & $39.9_{\pm0.2}$ & $42.4_{\pm0.2}$ & $43.0_{\pm0.2}$ & $39.9_{\pm0.2}$ & $42.7_{\pm0.2}$ & $\mathbf{47.8_{\pm0.2}}$ \\
\cmidrule(lr){2-9}

& InternVL3-2B
& $34.8_{\pm0.2}$ & $31.1_{\pm0.1}$ & $34.1_{\pm0.2}$ & \underline{$35.8_{\pm0.2}$} & $31.1_{\pm0.2}$ & $32.8_{\pm0.2}$ & $\mathbf{38.1_{\pm0.2}}$ \\

& InternVL3-8B
& $38.1_{\pm0.3}$ & $34.5_{\pm0.4}$ & $36.1_{\pm0.3}$ & \underline{$38.6_{\pm0.4}$} & $34.3_{\pm0.3}$ & $35.9_{\pm0.2}$ & $\mathbf{42.2_{\pm0.4}}$ \\

& InternVL3-14B
& $35.8_{\pm0.2}$ & $32.8_{\pm0.1}$ & $34.8_{\pm0.2}$ & \underline{$38.1_{\pm0.2}$} & $34.1_{\pm0.2}$ & $34.8_{\pm0.2}$ & $\mathbf{45.1_{\pm0.2}}$ \\

& Qwen3.5-VL-2B
& $33.2_{\pm0.2}$ & $31.5_{\pm0.2}$ & $34.1_{\pm0.2}$ & \underline{$35.8_{\pm0.1}$} & $30.9_{\pm0.2}$ & $33.8_{\pm0.1}$ & $\mathbf{37.0_{\pm0.2}}$ \\

& Qwen3.5-VL-8B
& $32.9_{\pm0.1}$ & $30.9_{\pm0.3}$ & \underline{$33.8_{\pm0.2}$} & \underline{$36.1_{\pm0.2}$} & $32.1_{\pm0.2}$ & $33.1_{\pm0.1}$ & $\mathbf{42.1_{\pm0.2}}$ \\

& Qwen3.5-VL-27B
& $34.1_{\pm0.2}$ & $31.9_{\pm0.2}$ & $33.8_{\pm0.2}$ & \underline{$35.8_{\pm0.3}$} & $30.8_{\pm0.1}$ & $33.8_{\pm0.3}$ & $\mathbf{44.1_{\pm0.2}}$ \\

\midrule

\multirow{8}{*}{VRS-Bench-VQA}

& GeoChat-7B
& $65.3_{\pm0.2}$ & $61.3_{\pm0.2}$ & $63.5_{\pm0.2}$ & \underline{$65.9_{\pm0.2}$} & $61.3_{\pm0.2}$ & $64.1_{\pm0.2}$ & $\mathbf{69.9_{\pm0.2}}$ \\

& GeoLLaVA-8B
& \underline{$70.8_{\pm0.2}$} & $65.1_{\pm0.2}$ & $67.8_{\pm0.2}$ & $70.7_{\pm0.2}$ & $65.1_{\pm0.2}$ & $66.8_{\pm0.2}$ & $\mathbf{74.9_{\pm0.2}}$ \\
\cmidrule(lr){2-9}

& InternVL3-2B
& \underline{$65.4_{\pm0.3}$} & $60.9_{\pm0.4}$ & \underline{$65.4_{\pm0.3}$} & $65.1_{\pm0.4}$ & $61.0_{\pm0.3}$ & $63.0_{\pm0.4}$ & $\mathbf{68.5_{\pm0.3}}$ \\

& InternVL3-8B
& $68.3_{\pm0.4}$ & $65.4_{\pm0.3}$ & $67.4_{\pm0.1}$ & \underline{$70.2_{\pm0.1}$} & $65.8_{\pm0.3}$ & $63.1_{\pm0.4}$ & $\mathbf{72.8_{\pm0.2}}$ \\

& InternVL3-14B
& $70.2_{\pm0.3}$ & $66.1_{\pm0.4}$ & $65.8_{\pm0.3}$ & \underline{$70.9_{\pm0.3}$} & $65.1_{\pm0.3}$ & $65.2_{\pm0.3}$ & $\mathbf{75.2_{\pm0.4}}$ \\

& Qwen3.5-VL-2B
& \underline{$64.8_{\pm0.1}$} & $60.2_{\pm0.2}$ & $64.7_{\pm0.1}$ & $64.7_{\pm0.2}$ & $61.9_{\pm0.2}$ & $61.1_{\pm0.2}$ & $\mathbf{66.1_{\pm0.3}}$ \\

& Qwen3.5-VL-8B
& $67.8_{\pm0.2}$ & $64.8_{\pm0.1}$ & $67.8_{\pm0.2}$ & \underline{$69.9_{\pm0.2}$} & $62.8_{\pm0.2}$ & $67.1_{\pm0.2}$ & $\mathbf{73.0_{\pm0.1}}$ \\

& Qwen3.5-VL-27B
& $69.9_{\pm0.2}$ & $65.8_{\pm0.2}$ & $67.1_{\pm0.2}$ & \underline{$70.9_{\pm0.1}$} & $65.8_{\pm0.2}$ & $65.8_{\pm0.1}$ & $\mathbf{74.1_{\pm0.2}}$ \\

\bottomrule
\end{tabular}
\end{table*}

\subsection{Analyses}

\begin{table*}[h]
\centering
\caption{Per-task accuracy ($\%$) on UrBench-MCQ with tasks grouped into vision and language branches.}
\label{tab:per_task_results}
\small
\setlength{\tabcolsep}{2.65pt}
\renewcommand{\arraystretch}{1.12}
\begin{tabular}{l|ccccc|ccccccccc|ccc}
\toprule
\multirow{2}{*}{Model}
&
\multicolumn{5}{c|}{\textbf{Vision-oriented (V) Tasks}}
&
\multicolumn{9}{c|}{\textbf{Language-oriented (L) Tasks}}
&
\multicolumn{3}{c}{\textbf{Average}}\\
\cmidrule(lr){2-6}\cmidrule(lr){7-15}\cmidrule(l){16-18}
&
Cam.&Traffic&Road&Ori.&Match
&
City&Count&Img.&Attr.&Ground&Role&Comp.&Scene&VPR
&
V &L &All\\
\midrule

Qwen3.5-VL-27B (Joint)
&41.2&68.1&66.7&30.1&59.8
&70.2&78.1&45.3&49.9&70.4&71.1&85.7&76.1&85.2
&53.2&70.2&65.1\\

Qwen3.5-VL-27B (G$^2$DA)
&45.0&72.0&72.5&32.5&62.6
&70.0&83.8&45.0&52.5&72.5&75.0&90.0&85.0&88.9
&56.9&73.6&68.7\\

\hline

GeoLLaVA-8B (Joint)
&38.2 & 73.1 & 50.1 & 24.9 & 44.3
& 54.8 & 49.1 & 40.2 & 46.2 & 62.4 & 63.3 & 54.5 & 53.2 & 70.2
& 46.1 & 54.9 &51.7\\

GeoLLaVA-8B (G$^2$DA)
&40.0&75.0&56.7&25.0&48.2
&55.0&54.1&41.0&49.0&65.0&67.5&58.7&62.5&74.0
&49.0&58.5&54.8\\

\bottomrule
\end{tabular}
\end{table*}

\paragraph{Existing multi-task optimization methods show limited transferability to Geo-VLMs.} From Table~\ref{tab:benchmark}, none of the representative multi-task optimization methods consistently outperform joint training with random task shuffling. PCGrad and CAGrad incur the largest performance degradation, while GradNorm and FAMO exhibit comparatively smaller drops, although neither surpasses the joint-training baseline. This suggests that optimization strategies developed for conventional multi-task learning do not readily transfer to Geo-VLMs. One possible explanation is that the substantially higher task heterogeneity  limits the effectiveness of generic gradient-based optimization. In particular, explicitly modifying gradient directions, as in PCGrad and CAGrad, may become unreliable when tasks differ significantly, while gradient reweighting methods such as GradNorm and FAMO are less disruptive but remain insufficient. 

In contrast, STGU achieves more competitive performance by incorporating task scheduling rather than directly modifying gradient directions, suggesting that organizing heterogeneous tasks may be a more effective alternative to explicit gradient manipulation. G$^2$DA further extends this optimization paradigm by first performing gradient-guided cross-modal task partitioning and then constructing modality-specific curricula within each branch. This hierarchical organization reduces both cross-modal interference and intra-branch task conflicts, consistently outperforming STGU across all benchmarks and model architectures. Together, these results suggest that modality-aware optimization better leverages task heterogeneity for Geo-VLM adaptation.

\paragraph{Compatibility across model architectures.}

G$^2$DA consistently improves performance across both general-purpose and remote sensing-specific vision-language models, demonstrating good architectural compatibility. The gains are particularly pronounced on InternVL3 and GeoLLaVA, both of which adopt relatively decoupled vision-language architectures that better align with the proposed cross-modal optimization strategy. In contrast, improvements on Qwen3.5-VL are generally smaller but remain consistent across all model scales, indicating that G$^2$DA does not depend on a specific backbone design. Furthermore, the consistent improvements observed on GeoChat and GeoLLaVA suggest that the proposed optimization framework generalizes beyond general-purpose VLMs to models specifically designed for remote sensing applications.

\paragraph{Benchmark-dependent behavior.} The performance gains of G$^2$DA vary across benchmarks, likely reflecting differences in benchmark characteristics. The largest improvements are observed on XLRS-Bench-Lite, which contains higher-resolution imagery and more fine-grained reasoning tasks. By comparison, VRS-Bench shows more moderate gains, possibly because its lower-resolution imagery and recognition-oriented tasks present a comparatively simpler optimization problem. UrBench also exhibits substantial improvements, likely due to the optimization heterogeneity arising from diverse street-view, cross-view, and remote-sensing tasks. These observations suggest that G$^2$DA is more beneficial when visual representation learning or task heterogeneity becomes more challenging.



\paragraph{Task-aware analysis.}

We further examine per-task performance on UrBench-MCQ in Table~\ref{tab:per_task_results} by grouping tasks according to the vision- and language-oriented partition. Compared with joint training, G$^2$DA consistently improves nearly all tasks, with only marginal performance drops on Qwen3.5-VL-27B for City Retrieval ($-0.2$) and Image Retrieval ($-0.3$), possibly because these retrieval tasks are particularly challenging under limited training data and rely more heavily on the pretrained language model's semantic representations~\cite{zhou2025urbench}. The improvements are well balanced across both modalities, with Qwen3.5-VL-27B achieving gains of +3.7 and +3.4 percentage points on vision- and language-oriented tasks, respectively, and GeoLLaVA-8B showing comparable improvements of +2.9 and +3.6 percentage points. This balanced improvement suggests that G$^2$DA effectively mitigates cross-modal optimization interference without introducing modality bias. Furthermore, the largest gains are consistently observed on reasoning-intensive tasks, including Scene Recognition, Road Understanding, and Counting, which exhibit the strongest gradient conflicts before decoupling (Fig.~\ref{fig:gradient_similarity}) and remain among the most conflicting tasks even after decoupling (Fig.~\ref{fig:local_matrices}), indicating that G$^2$DA is particularly beneficial for tasks with persistent optimization conflicts.

\paragraph{Gradient interaction after training.}
We further evaluate gradient conflict alleviation after training. We measure the conflict rate as the proportion of task pairs with negative cosine similarity,
\begin{equation}
R_{\mathrm{conflict}}
=
\frac{2}{N(N-1)}
\sum_{i<j}
\mathbb{I}(S_{ij}<0),
\label{eq:conflict_rate}
\end{equation}
where $\mathbb{I}(\cdot)$ is the indicator function.
A lower conflict rate indicates fewer conflicting optimization directions across tasks. The results are listed in Table~\ref{tab:gradient_statistics}. Across all benchmarks, G$^2$DA consistently increases the average pairwise gradient cosine similarity while reducing the proportion of conflicting task pairs, although the degree of improvement varies. Compared with joint training, gradient conflicts are reduced by up to 6.0 percentage points, while gradient alignment increases by 5.1-13.1$\%$. The largest improvements are observed on UrBench-MCQ, likely due to its diverse cross-view and multi-view tasks with greater optimization heterogeneity. In contrast, VRS-Bench-VQA shows lower gradient conflicts and smaller gains, likely because its predominantly recognition-oriented tasks and lower-resolution imagery yield a more compatible optimization landscape, leaving less room for improvement.

\begin{table}[t]
\centering
\caption{Gradient interaction statistics after training. Higher cosine similarity and lower conflict rate ($\%$) indicate less gradient interference.}
\label{tab:gradient_statistics}
\small
\setlength{\tabcolsep}{3.5pt}
\renewcommand{\arraystretch}{1.0}
\begin{tabular}{lcc|cc|cc}
\toprule
& \multicolumn{2}{c|}{UrBench-MCQ}
& \multicolumn{2}{c|}{XLRS-Bench-Lite}
& \multicolumn{2}{c}{VRS-Bench-VQA} \\
Method
& Cos.$\uparrow$ & Conf.$\downarrow$
& Cos.$\uparrow$ & Conf.$\downarrow$
& Cos.$\uparrow$ & Conf.$\downarrow$ \\
\midrule
Joint
& 0.0191 & 47.3
& 0.0221 & 43.0
& 0.0236 & 41.2 \\
G$^2$DA
& \textbf{0.0216} & \textbf{41.3}
& \textbf{0.0240} & \textbf{41.1}
& \textbf{0.0248} & \textbf{40.5} \\
\bottomrule
\end{tabular}
\end{table}

\subsection{Ablation Study}

We conduct an ablation study to evaluate the contribution of the three key components in G$^2$DA: gradient-guided cross-modal Decoupling (D), gradient-aware Curriculum (C), and bidirectional Rehearsal (R). Table~\ref{tab:ablation} summarizes the results on two benchmarks using two representative model architectures. Gradient-guided task decoupling consistently provides the largest individual improvement over joint multi-task training, with gains of +4.1/+1.8 pp on InternVL3-8B and +2.6/+1.9 pp on GeoLLaVA-8B for UrBench/XLRS, respectively. This demonstrates that separating tasks according to their dominant optimization modality effectively alleviates cross-modal optimization interference. Building upon decoupling, gradient-aware curriculum scheduling further improves performance on the more challenging XLRS benchmark (+0.8 and +0.5 pp for InternVL3-8B and GeoLLaVA-8B, respectively), indicating that ordering tasks according to gradient compatibility is particularly beneficial under stronger optimization heterogeneity. On UrBench, however, curriculum scheduling yields a slight performance drop compared with decoupling alone (56.0$\rightarrow$54.9 and 53.3$\rightarrow$52.1), suggesting that the benefit of gradient-aware ordering may be partially offset by the recency effects associated with sequential training. Finally, incorporating bidirectional rehearsal consistently recovers and further improves performance, achieving the best results across all benchmark--model combinations. These results demonstrate that the three components are complementary: decoupling mitigates cross-modal interference, curriculum exploits intra-modal gradient compatibility, and bidirectional rehearsal alleviates the recency effects introduced by sequential optimization.

\begin{table}[h]
\centering
\caption{Component ablation of G$^2$DA (accuracy, $\%$).}
\label{tab:ablation}
\small
\setlength{\tabcolsep}{2pt}
\renewcommand{\arraystretch}{1.}
\begin{tabular}{lccc|cc|cc}
\toprule
\multirow{2}{*}{Method}
& \multicolumn{3}{c|}{Components}
& \multicolumn{2}{c|}{InternVL3-8B}
& \multicolumn{2}{c}{GeoLLaVA-8B} \\
\cmidrule(lr){2-4}
\cmidrule(lr){5-6}
\cmidrule(l){7-8}
& D & C & R
& UrBench
& XLRS
& UrBench
& XLRS \\
\midrule
Joint                &  &  &  & 51.9 & 38.1 & 50.7 & 43.8 \\
+ Decoupling         & \checkmark &  &  & 56.0 & 39.9 & 53.3 & 45.7 \\
+ Curriculum         & \checkmark & \checkmark &  & 54.9 & 40.7 & 52.1 & 46.2 \\
G$^2$DA              & \checkmark & \checkmark & \checkmark & 57.8 & 42.2 & 54.8 & 47.8 \\
\bottomrule
\end{tabular}
\end{table}








\section{Conclusion}

This paper investigated the optimization challenges of multi-task Geo-VLM learning from a gradient-centric perspective. Our empirical analysis revealed that heterogeneous geospatial tasks exhibit substantial optimization conflicts, including inter-task gradient interference, distinct cross-modal optimization preferences, and persistent intra-encoder gradient heterogeneity. To address these challenges, we proposed G$^2$DA, which integrates gradient-guided cross-modal decoupling, branch-specific curriculum scheduling, and bidirectional rehearsal into a unified optimization framework. Extensive experiments on three Geo-VLM benchmarks using six general-purpose VLMs and two representative Geo-VLMs demonstrate that G$^2$DA consistently outperforms representative multi-task optimization baselines across all 24 benchmark--model settings under identical optimization budgets. These results indicate that explicitly organizing optimization according to task-level gradient characteristics effectively alleviates optimization interference and provides an effective optimization paradigm for multi-task Geo-VLM learning.

\bibliography{aaai2027}



\end{document}